\documentclass{article}

\usepackage{PRIMEarxiv}
\usepackage{eccvabbrv}

\usepackage[utf8]{inputenc} 
\usepackage[T1]{fontenc}    
\usepackage{hyperref}       
\usepackage{url}            
\usepackage{booktabs}       
\usepackage{amsfonts}       
\usepackage{nicefrac}       
\usepackage{microtype}      
\usepackage{lipsum}
\usepackage{fancyhdr}       
\usepackage{graphicx}       
\graphicspath{{media/}}     

\usepackage{amsmath}
\usepackage{pifont}
\usepackage{makecell}
\usepackage{multirow}
\usepackage{comment}
\usepackage{multicol}
\usepackage{pifont}
\usepackage{makecell}
\usepackage{algorithm}
\usepackage{algorithmic}
\usepackage{array}
\usepackage[dvipsnames]{xcolor}

\usepackage{colortbl}
\usepackage{wrapfig}

\usepackage[accsupp]{axessibility}  

\usepackage{authblk}
\usepackage{caption}

\usepackage{paralist}

\newcommand{\myparagraph}[1]{\noindent{\bf #1.}}

\newcommand{\calT}{\mathcal{T}}
\newcommand{\calL}{\mathcal{LP}}

\newcommand{\calI}{\mathcal{I}}

\newcommand{\real}{\mathbb{R}}

\newcommand{\CUT}[1]{}

\title{Fine-grained CLIP fine-tuning with self-annotated region alignment
}

\author[1,2]{Chenyang~Zhao}
\author[1]{Wei~Lin}
\author[2]{Janet~H.~Hsiao}
\author[1]{Antoni~B.~Chan}

\affil[1]{Department of Computer Science, City University of Hong Kong}
\affil[2]{Division of Social Science, Hong Kong University of Science \& Technology}

\begin{document}
\maketitle

\begin{abstract}
Contrastive Language-Image Pre-training (CLIP) has been shown to have limitations in its fine-grained dense feature representation, due to its pre-training focusing on matching the whole image to a text description.
Considering the large data and computational burden in pre-training a vision-language model from scratch, a series of works aim to enhance the fine-grained ability of CLIP through a fine-tuning scheme. 
However, existing works suffer from a variety of limitations: additional region annotations are usually required, which limits the semantic diversity due to the predefined categories and leads to a large effort to process the training data; and they usually sacrifice CLIP's original ability for global visual representation. 
To bypass these limitations, we propose SFF-CLIP (Self-annotated Fine-grained Fine-tuning for CLIP), which only uses image-text pairs as input to boost the fine-grained representation ability in the CLIP fine-tuning, while maintaining the global visual-semantic consistency. 
Concretely, a run-time region-phrase alignment scheme is designed, which obtains concept phrases from the input sentence, and aligns them with corresponding extracted region-based features using text-specific heat maps.
Extensive experiments demonstrate that SFF-CLIP leads to significant performance improvements on fine-grained dense feature representation, as well as maintaining the performance of the original CLIP on image-level tasks. 
Code will be released later. 
\end{abstract}


\section{Introduction}
\label{sec:intro}
Contrastive Language-Image Pre-training (CLIP) \cite{radford2021learning} aligns global image and text embeddings within a unified latent space by adopting a dual-encoder architecture and conducting matching on large-scale, noisy image-text pairs. It has become a foundational vision-language model (VLM) for representation learning, and achieves remarkable success on image-level tasks \cite{changpinyo2021conceptual, cha2022domain, luo2022clip4clip}, such as image classification and image-text retrieval. 

However, CLIP exhibits notable limitations in comprehending fine-grained details, such as poor region recognition when using its dense features, due to the pre-training focusing on matching the whole image (via the $[cls]$ token) to a text description. Thus, the model struggles to extract meaningful region-level representations from its dense visual features for grounding textual concepts.
This shortcoming limits the performance of CLIP on the downstream tasks that require region-aware ability. For example, in dense prediction tasks, e.g., object detection and segmentation, the CLIP model is usually utilized as a classifier \cite{xu2021simple, liang2023open} or the teacher in distillation \cite{gu2021open, du2022learning} to process cropped object patches to obtain region features. Some works \cite{kuo2023fvlm, wu2023cora, yu2024fcclip} adopt the frozen CLIP model as the backbone to produce spatial feature maps, but they all choose the CNN-based CLIP, which can preserve more position information than the vision transformer (ViT-based) architecture.

\begin{figure*}[t]
	\centering
	\vspace{-0.2cm}
	\begin{center}
		\includegraphics[width=0.98 \textwidth]{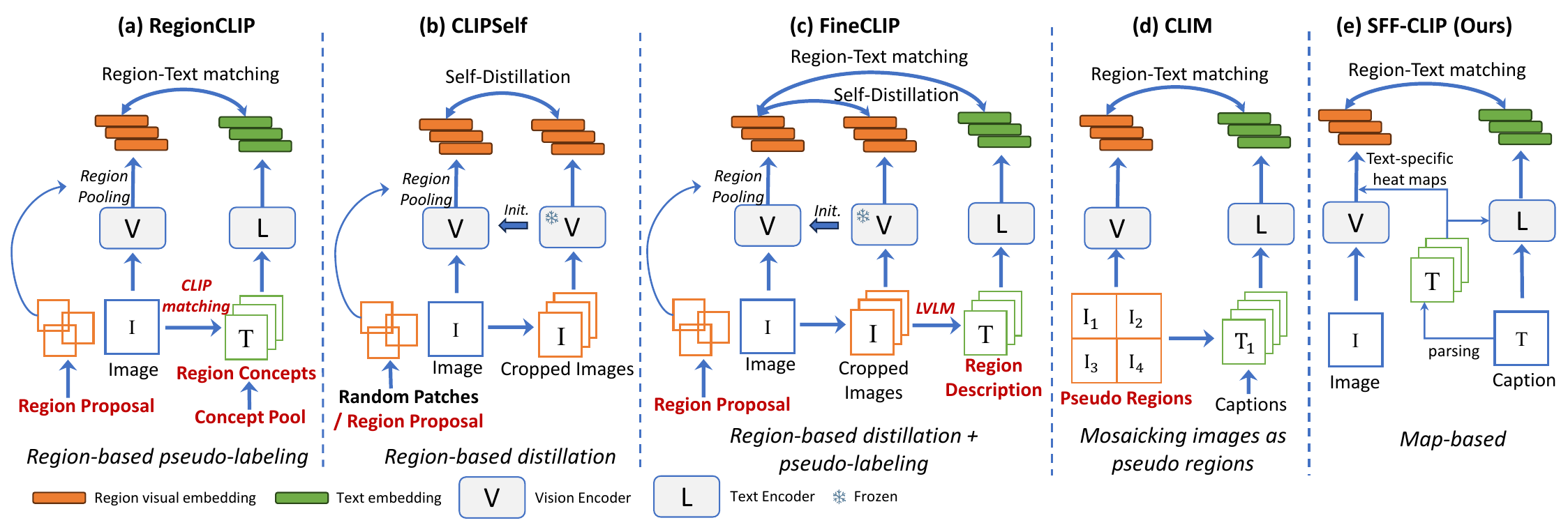}
	\end{center}
	\vspace{-0.2cm}
	\caption{Comparison of different fine-grained alignment methods for CLIP. The required preparations for region annotation are marked with bold red texts. Our method eliminates the constraints of pre-defined category limitation, and the large effort required for generating region proposals and their corresponding descriptions.
	}
	\vspace{-0.2cm}
	\label{fig:figure1}
\end{figure*}

Since the resource demand for pretraining a fine-grained VLM is staggering, such as billions of data samples \cite{tschannen2025siglip, xiefg2025fgclip, bica2024improving}, a series of recent works have focused on advancing the vision-language alignment of local image regions via fine-tuning of CLIP. Considering the impressive generalization capabilities of CLIP models, particularly the powerful ViT-based variants, the fine-grained fine-tuning of CLIP is carried out to enable applications on dense prediction tasks, such as open-vocabulary object detection and image segmentation.
Since no region-text annotations are provided in the image-text pair training data, the previous fine-tuning works enhance the fine-grained alignment via: 1) \emph{region-based pseudo-labeling}, 2) \emph{region-based distillation}, 3) \emph{mosaicking images as pseudo regions}, or the \emph{combination} thereof.
In the first category, 
\cite{wang2023position, zhong2022regionclip, li2025densevlm} generate region proposals by coarsely cropping patches or adopting RPN \cite{ren2015faster}, and then use CLIP or other powerful VLMs to retrieve region labels from a pool of concepts (\eg, Fig.~\ref{fig:figure1}a). However, the range of concepts is limited by the number of pre-defined categories, which could hinder downstream applications like open-vocabulary object detection. 
The recent work FineCLIP \cite{jing2024fineclip} (Fig.~\ref{fig:figure1}c) adopts large VLMs like BLIP2 \cite{li2023blip} to generate detailed sentences to describe region proposals. 
However, carefully preparing the region annotations in these methods inevitably costs significant extra time and space, which cannot be neglected when increasing the scale of training data. 
For the 2nd category, 
CLIPSelf \cite{wu2023clipself} (Fig.~\ref{fig:figure1}b) and FineCLIP \cite{jing2024fineclip} (Fig.~\ref{fig:figure1}c) use distillation to transfer the global features of the cropped regions (either random or proposals) to extracted dense features -- however, to obtain superior distillation performance, preprocessing of the generated region proposals by a well-trained detector is still required. 
Mosaic augmentation is adopted to make pseudo regions in \cite{yeo2025atas, wu2024clim} (\eg, Fig.~\ref{fig:figure1}d), which produces finer matching than image-level while still coarser than region alignment.
Moreover, most of these methods sacrifice, and some even ignore \cite{wu2023clipself,wu2024clim,yeo2025atas,li2025densevlm},  the global representation ability of CLIP when training focuses on region knowledge, which causes severe performance damage on image-level tasks.

In this work, we propose SFF-CLIP, a novel self-annotated fine-grained fine-tuning method for CLIP, which aims to boost the region awareness of CLIP while maintaining the global visual-semantic consistency, \emph{without} requiring extra preparation of region proposals or their annotations.  
As with the original CLIP pre-training, image-text pairs are taken as inputs for fine-tuning. We propose a self-annotated fine-grained alignment scheme, where we first extract phrases from the input sentence and generate the corresponding region importance on the image based on text-specific heat maps. The heat maps are used to aggregate region features for each phrase, and then the region-phrase feature pairs are aligned by the matching loss, whose weight is based on the degree of phrase matching. 
To help maintain the whole image representation ability of the original CLIP, the contrastive learning objective for image-text pairs is included in the loss function through weighted momentum term. 
As a result, our SFF-CLIP locally and globally aligns the visual and semantic features in the same representation space. 
Through comprehensive experiments, we show that SFF-CLIP significantly improves the fine-grained understanding of CLIP, surpasses previous state-of-the-arts on both the dense prediction benchmarks, including open-vocabulary object detection and segmentation, while also maintaining the global representation of the original model in the image-level task. 

We summarize our contributions as follows: 
\begin{compactenum}
\item We propose SFF-CLIP, a self-annotated fine-grained alignment strategy for finetuning CLIP, which uses a run-time
scheme that generates region knowledge from input image-text pairs to achieve better fine-grained understanding.

\item SFF-CLIP eliminates the need for extra region-based training data, such as pre-defined categories, or generating region proposals and their corresponding descriptions.

\item SFF-CLIP successfully maintains the global representation capability of CLIP while improving the region awareness. 

\item Extensive experiments on dense prediction and image-level tasks show that the proposed SFF-CLIP consistently outperforms previous methods.
\end{compactenum}

\section{Related Work}
\label{sec:relatedwork}

\subsection{Fine-grained understanding in VLMs}
Although CLIP \cite{radford2021learning} and subsequent VLMs \cite{zhou2022learning,yu2022coca,li2022blip} exhibit strong representation capabilities and exceptional generalizability,
the image-level training, which matches an image as a whole to a text description, has been shown to be deficient in fine-grained understanding and alignment between image regions and text \cite{zhong2022regionclip,wang2023position, kim2023region,wu2023clipself}. This shortcoming limits their applications on tasks that require region-aware abilities, \eg, dense prediction tasks.

To mitigate this issue, some works build strong fine-grained VLMs by pre-training with patch-token embedding alignment \cite{bica2024improving}, or region-caption matching \cite{tschannen2025siglip} utilizing well-trained open-vocabulary detectors for assistance. FG-CLIP \cite{xiefg2025fgclip} constructs a comprehensive dataset with billions of region-specific annotations for region-aware training for CLIP. Due to the heavy cost of using large-scale data, some recent works propose to enhance the fine-grained representation by fine-tuning CLIP with region-based pseudo-labeling or region-based distillation.
RegionCLIP \cite{zhong2022regionclip} adopts RPN \cite{ren2015faster} object proposals, while PTP \cite{wang2023position} coarsely crops patches, and then they both use CLIP as a classifier to obtain region labels from a large pre-defined pool of concepts.
DenseVLM \cite{li2025densevlm} uses a powerful VLM to retrieve categories (from a category set) for randomly cropped regions. 
The recent work FineCLIP \cite{jing2024fineclip} adopts large VLMs like BLIP2 \cite{li2023blip} to generate detailed sentences to describe region proposals. 
However, the semantic diversity is limited for those works with pre-defined categories or concepts, and these methods inevitably cost significant extra time and space to preprocess the region annotations, which cannot be neglected when increasing the training data. 
CLIPSelf \cite{wu2023clipself} and ATAS \cite{yeo2025atas} facilitate the transfer of the global features of regions to dense feature extraction by self-distillation -- CLIPSelf obtains the region proposals through the object detector, and ATAS uses mosaic augmentation to construct pseudo regions as in CLIM \cite{wu2024clim}. The distillation-based methods only focus on the image encoder training and ignore the text encoder. Furthermore, among these methods, \cite{li2025densevlm, wu2023clipself, yeo2025atas, wu2024clim} are also ignore the global image-text matching ability of CLIP.

Our paper proposes a self-annotated fine-grained fine-tuning framework for boosting the region-aware ability of CLIP that \emph{only uses image-text pairs as input and does not use external models}. Our approach circumvents the resource-consuming preparation of the region annotations and the limitation of pre-defined visual categories, while successfully preserving the global representation ability of CLIP.

\subsection{Open-vocabulary dense prediction}
Open-vocabulary dense prediction aims to identify visual regions of arbitrary categories as described by the text, and primarily comprises open-vocabulary object detection \cite{gu2021open, kuo2023open, minderer2022simple} and image segmentation \cite{ghiasi2022scaling,li2024cascade, liang2023open}. Recent open-vocabulary approaches have leveraged the strong representation and generalizability displayed by powerful pre-trained VLMs, which are exploited to identify novel objects. Due to CLIP lacking precise local vision-language alignment, CLIP is usually adopted as a classifier \cite{xu2021simple, liang2023open} or the teacher in distillation \cite{gu2021open, du2022learning} to process cropped region proposals. Several works \cite{kuo2023fvlm, wu2023cora, yu2024fcclip, kuo2023open, ghiasi2022scaling} adopt frozen CLIP encoders as backbone in the detectors to generate visual features, but most of them select the CNN-based CLIP, which can preserve more position information than the vision transformer (ViT-based) architecture. Although recent studies \cite{zhong2022regionclip,jing2024fineclip,wu2023clipself,li2025densevlm, wu2024clim} fine-tuning CLIP for fine-grained understanding, they are constrained by the limitations when generating dense annotations as mentioned above.

\subsection{Visual explanation of VLMs}
Visual explanation methods \cite{zeiler2014visualizing,selvaraju2017grad,ribeiro2016should, petsiuk2018rise, zhao2023odam} in computer vision have been developed 
for visually interpreting a specific prediction of the model by generating a heat map that indicates the spatial feature importance.
Recently,  works \cite{qiang2022attcat,xie2022vit,yu2023x,zhao2024gradient} have proposed visual explanations for the Transformer architecture. 
In our designed run-time self-annotation scheme for region-phrase pairs, we produce text-specific heat maps using Grad-ECLIP \cite{zhao2024gradient} as reference, which is a state-of-the-art visual explanation method for VLMs, such as CLIP. The heat map generation process is high-speed and can be easily plug-in the training of proposed fine-grained alignment.

\section{Method}
\label{sec:method}

\begin{figure*}[t]
	\centering
	\vspace{-0.1cm}
	\begin{center}
		\includegraphics[width=0.98 \textwidth]{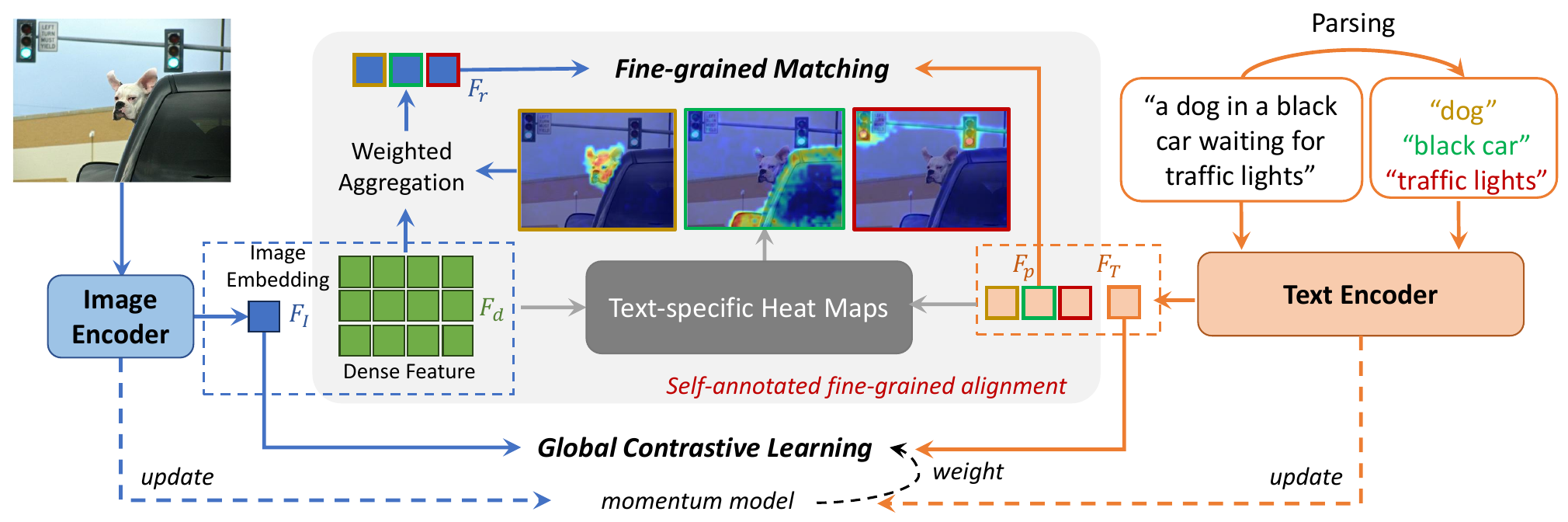}
	\end{center}
	\vspace{-0.2cm}
	\caption{Overview of the proposed SFF-CLIP. Multiple phrases or words representing objects (e.g., ``dog'', ``black car'' and ``traffic lights'') are separated out by parsing the input caption. 
	By generating text-specific activation maps, object-specific region feature embeddings ($F_{r}$) are obtained through weighted aggregation of the image dense feature ($F_{d}$). 
    Then the image region features ($F_{r}$) and the corresponding phrase features ($F_{p}$) are aligned in the training scheme, together with the global contrastive learning.
	}
	
	\vspace{-0.2cm}
	\label{fig:fine_grained_clip}
\end{figure*}

The aim of SFF-CLIP is to realize region-language alignment while maintaining image-level representation in the fine-tuning of CLIP, but without the requirements of generating dense data annotations, including pre-defined categories, region proposals, or region descriptions.
To achieve this, the vision-language alignment at region and image levels are unified in our training method, employing our proposed run-time self-annotated fine-grained alignment and CLIP's original global contrastive learning. 
The framework of our proposed SFF-CLIP is shown in Fig.~\ref{fig:fine_grained_clip}, where the model has the same architecture as the original CLIP.
We first present brief preliminaries about CLIP (\S\ref{sec:pre_clip}), then introduce fine-grained alignment (\S\ref{sec:local_align}) and global alignment (\S\ref{sec:global_align}) in detail.  

\subsection{Preliminaries}
\label{sec:pre_clip}

CLIP learns visual and language representations from large-scale 
web-curated image-text pairs. It consists of an image encoder $\calI\left(\cdot\right)$ and a text encoder $\calT\left(\cdot\right)$, which are jointly trained to extract image and text feature embeddings in a unified representation space. Given image-text pair $(I,T)$, the matching score between their extracted 
image features $F_{I}\in \mathbb{R}^{D}$ and text features $F_{T}\in \mathbb{R}^{D}$ (both row vectors) is:
\begin{align}
	S(F_{I}, F_{T}) = \cos(F_{I}, F_{T}) = \tfrac{F_{I}F_{T}^\mathsf{T}}{\left \| F_{I} \right \| \left \| F_{T} \right \| } .
	\label{eq:cos_similarity}
\end{align}
The model is trained using contrastive learning on the matching scores, regarding the ground-truth image-text pairs as positive samples and other mismatched pairs as negatives. 
For ViT-based encoders, the image feature is $F_{I}=\calL(x_{cls})$, where $\calL$ denotes linear projections, and $x_{cls}$ is the feature vector from the $[cls]$ token. Thus, except for the class token, all the final layer features of the other tokens (image patch tokens) are not used during contrastive learning of CLIP. Since only the class token feature is explicitly optimized during training, the local patch features 
exhibit weak representation ability for localized matching of semantic content. 

\subsection{Self-annotated fine-grained alignment}
\label{sec:local_align}

To  realize fine-grained alignment, we propose a matching loss based on extracting the dense feature map from the image and performing fine-grained feature matching of regions to the corresponding text phrase. 
We dynamically generate self-annotated region-phrase pairs from the input image-text pairs during the training process, which we denote as \emph{run-time} self-annotation, instead of off-line preparation with detectors or pre-defined categories.
Note that our run-time self-annotation is based on the concepts learned by the original CLIP, which makes it more flexible than using pre-defined detectors, since it is not limited to a pre-defined pool of visual categories.

\subsubsection{Image dense feature}
Following \cite{zhou2022extract, wu2023clipself}
we extract a dense feature map of the input image from a ViT-based encoder by slightly modifying the last transformer layer to keep the projection and norm layers, and discard the self-attention. This modification is experimentally shown to be capable of preserving more spatial detailed features in the output token embeddings \cite{zhou2022extract}.
Specifically, in the last transformer layer, with the input $x=(x_{cls}, x_{1},...,x_{h\times w})$ comprising a $[cls]$ embedding and $h \times w$ spatial token embeddings, the
layer's output is obtained as $v_{i}=\calL(x_i)$, where $v_{i}\in\real^C$ represent the \textit{value} embeddings at spatial location $i$, with $C$ as their channel dimension.
Then, the $[cls]$ embedding is removed and the final spatial token embeddings are reshaped into an $h \times w$ dense feature map $F_{d}$, from which we extract fine-grained representations for specific image regions.

\subsubsection{Run-time region-phrase self-annotation}
\label{sec:method_annotation}
We propose a run-time scheme to automatically obtain region-phrase pairs from the input image-text pairs. Note that our method does not require any manually annotated or model-generated region proposals or their corresponding label annotations. It is also completely encapsulated -- it does not require any external vision models (e.g., other VLMs).

\textbf{Phrase extraction.} For the caption $T$ in each image-text pair $(I, T)$, we use the Natural Language Toolkit (NLTK) \cite{bird2009natural} to parse and extract the phrases that contain object concepts, by setting the separation and selection rules as ``adjective + noun''. For the example in Fig.~\ref{fig:fine_grained_clip}, the parsing results are ``dog'', ``black car'', and ``traffic lights'' with the input text ``a dog in a black car waiting for traffic lights''. 
Then, for these extracted words or phrases $\lbrace p_{t} \rbrace_{t=1}^n$, where $n$ is the 
number of extracted phrases, 
textual descriptions are generated using prompt-template strategy \cite{radford2021learning}: ``This is a photo of the $p_{t}$.''
These descriptions are sent to the text encoder, resulting in a set of phrase embeddings $\{F_{p_{t}}\}_t$. 

\textbf{Text-specific heat map.} The region embedding $F_{r_t}$ corresponding to each phrase embedding $F_{p_t}$ is calculated using a text-specific heat maps based on explainable AI (XAI) methods. These heat maps have been proposed to visually explain a specific prediction of the model by showing the spatial feature importance with a heat map. The local regions with relatively high values are interpreted as being important for generating the current output \cite{selvaraju2017grad, zhao2024gradient,zhao2023odam}. Referring to Grad-ECLIP \cite{zhao2024gradient}, a high-speed and easy plug-in gradient-based heat map method designed for ViT-based VLMs, we calculate the spatial importance heat map corresponding to each extracted phrase. 
Specifically, the phrase embeddings $F_{p_{t}}$ are used to calculate the cosine similarities with the image embedding $F_I$, and each similarity $\cos(F_I, F_{p_t})$ is regarded as a target to be interpreted. 
Following \cite{zhao2024gradient}, the gradient ($g_c$) for each similarity w.r.t. the output of the last layer is used as channel weights to aggregate the feature maps $v$, and a loosened attention map technique 
is adopted to calculate the spatial weight $u_{i}=\Phi(\cos(q_{cls}, k_{i}))$, where $\Phi(\cdot)$ is min-max normalization.
 Finally, the heat map for a specific phrase $p_t$ is: 
\begin{align}
\footnotesize	
 H_{ti} = \textstyle\sum\nolimits_{c} g_c \cdot u_i \cdot v_i, 
\end{align} 
where $\sum_c$ is the channel-sum operator.
As shown in Fig.~\ref{fig:fine_grained_clip}, the heat map $H_t$ can localize the corresponding concept by revealing the important spatial locations for matching with the specific phrase.
The heat maps are adopted as weights for aggregating the image dense feature $F_{d}$, resulting in the region embedding for the phrase: $F_{r_{t}} = \sum_{hw} H_{t} \cdot F_{d}$, where $\sum_{hw}$ is the sum operator over spatial coordinates. 

To investigate the influence of the XAI method used for fine-grained alignment, we conduct an ablation study with several other high-speed Transformer-applicable visual explanation approaches in the Supplemental, which demonstrates that the Grad-ECLIP is the most effective.

\subsubsection{Fine-grained matching}

Given the region-phrase embeddings, the positive pairs are made to be similar, while the negative pairs are separated, via the loss function:
\begin{eqnarray}
\lefteqn{L_{fg} = -\sum_{t}w_{t} \Big[( 1-S( F_{r_{t}}, F_{p_{t}}))^{2} \log  S( F_{r_{t}}, F_{p_{t}}) } \nonumber\\
		& +  \sum_{t' \neq t} S( F_{r_{t}}, F_{p_{t'}})^{2} \log ( 1- S( F_{r_{t}}, F_{p_{t'}}) )  \Big],
		\label{eq:local_loss}
\end{eqnarray}
where $S$ represents the cosine similarity function, and $t$, $t'$ means the $t$-th and $t'$-th phrase in the same batch.

\begin{figure}[tb]
	\centering
	\begin{center}
		\includegraphics[width=0.9 \textwidth]{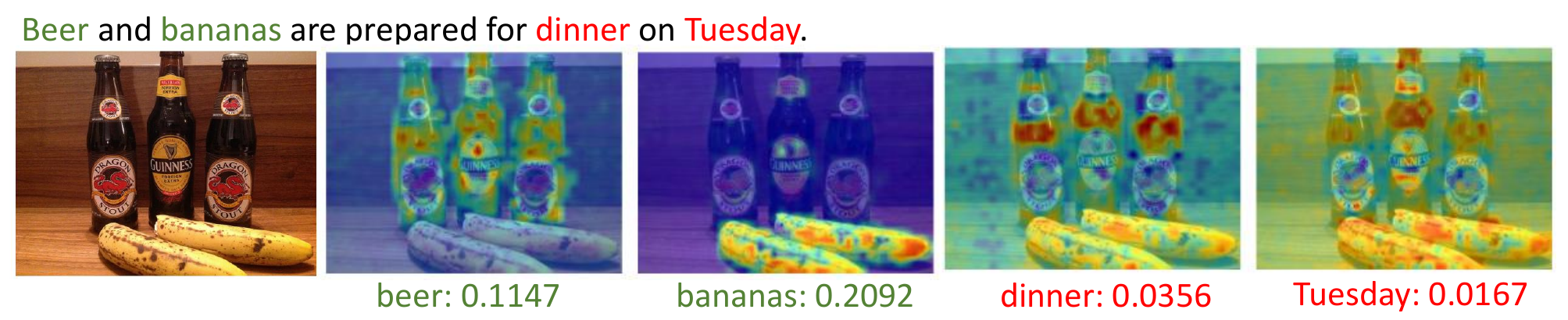}
	\end{center}
	\vspace{-0.2cm}
	\caption{An example of phrase alignment weight $w_{t}$, which is defined as the maximum value on the corresponding text-specific heat map. In this image-text pair, ``beer'' and ``banana'' are better matched with the image regions, as indicated by their higher weights $w_t$ than ``dinner'' and ``Tuesday''.
	}
	\vspace{-0.2cm}
	\label{fig:phrase_weights}
\end{figure}

A \textit{phrase-matching weight} $w_t$ is applied on the loss term for each region-phrase pair in order to diminish the potential impact of misaligned pairs, e.g., for extracted nouns that have no visual content in the image.
Specifically, we adopt  the highest response value on the phrase's heat map, 
i.e., $w_{t}=\max(H_{t})$, to evaluate the degree of matching between the phrase and the image. 
For example, in Figure~\ref{fig:phrase_weights}, ``dinner'' and ``Tuesday'' are extracted as nouns in the caption, while they are not visually matched with any of the image contents, and this can be reflected by the values on the heat map.
In practice, we set the maximum number of phrases in each sentence to  $N=5$. 
When the actual number of concepts $n$ is smaller than $N$, then blank phrases are added 
with weight $w_t=0$.

\subsection{Global contrastive learning with momentum}
\label{sec:global_align}
In standard CLIP training,
the input batch of image-text pairs $\lbrace (I_{b},T_{b}) \rbrace_{b=1}^B $ is passed to the image/text encoders and the model outputs the corresponding global image/text embeddings $\lbrace (F_{I_{b}}, F_{T_{b}} ) \rbrace_{b=1}^B$.  
The cosine similarity $S(F_{I}, F_{T})$ between image embedding $F_{I}$ and text embedding $F_{T}$ is calculated as in Eq.~\ref{eq:cos_similarity}, then the constrastive loss is applied to learn the global representations by maximizing the cosine similarities of the corresponding image and text embeddings, while minimizing the cosine similarities of other non-paired ones from both image and text sides, which is defined as:
\begin{equation}
\label{eqn:glossterm}
	\begin{aligned}
		L_{b}^{I2T} = -\log\tfrac{\exp\left( S(F_{I_{b}}, F_{T_{b}})/\tau \right) }{\sum_{b'=1}^{B} \exp\left( S(F_{I_{b}}, F_{T_{b'}})/\tau \right) }, 
		\ L_{b}^{T2I} = -\log\tfrac{\exp\left( S(F_{T_{b}}, F_{I_{b}})/\tau \right) }{\sum_{b'=1}^{B} \exp\left( S(F_{T_{b}}, F_{I_{b'}})/\tau \right) }, 
	\end{aligned}
\end{equation}
where $\tau$ is the trainable temperature parameter. 

To better maintain the image-level representation ability of the pre-trained CLIP  while enhancing its fine-grained alignment, we adopt a momentum CLIP model $\hat{M}$, which is initialized with the original CLIP model and then updated by the current trained model $M$ with a low momentum rate $\alpha=0.005$ after each epoch, \ie, $\hat{M} \leftarrow (1-\alpha)\hat{M} + \alpha M$. 
The updated model $\hat{M}$ is used to extract image and text embeddings $\lbrace (\hat{F}_{I_{b}}, \hat{F}_{T_{b}} ) \rbrace_{b=1}^B$, and weights are computed 
for each image-text pair, which measures how well they match:
{\small
\begin{equation}
		w_{I_{b}} = \gamma\tfrac{\exp\left( S(\hat{F}_{I_{b}}, \hat{F}_{T_{b}})/\tau \right)}{\sum_{b'=1}^{B} \exp\left( S(\hat{F}_{I_{b}}, 				\hat{F}_{T_{b'}})/\tau \right)} + (1-\gamma), \ \ 
		w_{T_{b}} = \gamma\tfrac{\exp\left( S(\hat{F}_{T_{b}}, \hat{F}_{I_{b}})/\tau \right)}{\sum_{b'=1}^{B} \exp\left( S(\hat{F}_{T_{b}}, 	\hat{F}_{I_{b'}})/\tau \right)} + (1-\gamma),
\end{equation}
}where parameter $\gamma$ 
controls the influence of match, 
which is set to $0.4$ by default in the experiments.
The contrastive loss then applies the weights to the loss terms in (\ref{eqn:glossterm}) to focus on preserving the well-matched image-text pairs,
\begin{equation}
	L_{contrastive} = \dfrac{1}{2B} \sum_{b=1}^{B} \left(w_{I_{b}}L_{b}^{I2T} + w_{T_{b}}L_{b}^{T2I} \right).
	\label{eq:global_loss}
\end{equation}
The final loss for training is obtained by 
by adding the fine-grained matching loss to the contrastive learning loss, 
$L=L_{fg} + L_{contrastive}$.

\section{Experiments}
\label{sec:exp}

We evaluate the proposed SFF-CLIP on both improving of fine-grained representation via zero-shot region classification, down-stream tasks of open-vocabulary detection and segmentation, and on maintaining of image-level representation via image-text retrieval task. We compare with the primary fine-grained fine-tuning methods, including RegionCLIP\cite{zhong2022regionclip}, CLIPSelf\cite{wu2023clipself}, FineCLIP\cite{jing2024fineclip}, DenseVLM\cite{li2025densevlm}, and CLIM\cite{li2025densevlm}. Moreover, we also compare with fine-tuning CLIP using only the global contrastive loss, \ie, without the fine-grained matching, which is a baseline for continuation of image-level training of CLIP (denoted as ``CLIP-g''). Finally, we carry out the ablation studies.

\subsection{Implementation details}
\label{sec:exp_setting}

For fairness, we re-trained all compared fine-grained fine-tuning methods, by initializing with the same pre-trained model and fine-tuning with the same training data, using the same input image size, which is the same as the CLIP pre-training, 224x224 for ViT-B/16 and 336x336 for ViT-L/14 by default.
Following previous works \cite{wu2023clipself, jing2024fineclip}, the experiments are conducted based on the pre-trained models from EVA-CLIP \cite{sun2023eva}.
Since some related works \cite{zhong2022regionclip, wu2023clipself, jing2024fineclip} that require region annotations open-sourced their region information files based on MS COCO train2017 set \cite{lin2014microsoft}, we conduct the fine-tuning on the same images with the captions provided by \cite{jing2024fineclip}. We mainly report the comparison results with other works on ViT-B/16, with the comparisons on ViT-L/14 provided in the Supplemental.
Two RTX 6000 Ada are used, and for our method, we use batch size 128, learning rate 1e-5, and weight decay of 0.1; for other methods, we use the training parameters provided in their codes.

\subsection{Comparisons on fine-grained representation}

\begin{table*}[t]
	\caption{Comparison of fine-grained dense representations via zero-shot classification on the ADE20K panoptic \cite{zhou2017scene} val set and COCO panoptic \cite{lin2014microsoft} val2017 set.
    We report the Top-1 and Top-5 mean accuracy on both object bounding boxes and panoptic masks. 
	The gray row is the baseline CLIP before fine-grained alignment fine-tuning. 
    Note that DenseVLM$^{\star}$ represents the results reported by the original paper \cite{li2025densevlm} for input 224x224, and $^{\dagger}$ represents the results testing with open-sourced fine-grained models pre-trained by large-scale data, shown in the brackets.
    Our method does not require preparation of region proposals (R.P.), region labels (R.L.), or predefined categories (P.C.) on the training data.
        }
	\label{tab:zero-shot}
	\centering
	\vspace{-0.2cm}    
	\begin{tabular}{lc|ccc|cc|cc||cc|cc@{}}
		\hline
		& &  &  &   & \multicolumn{4}{c||}{ADE20k} &  \multicolumn{4}{c}{MS COCO} \\
		&  &   &   &   & \multicolumn{2}{c}{Boxes} & \multicolumn{2}{c||}{Masks} & \multicolumn{2}{c}{Boxes} & \multicolumn{2}{c}{Masks}\\
		Method & Model  & R.P. &  R.L. & P.C.  & Top1 & Top5 & Top1 & Top5  & Top1 & Top5 & Top1 & Top5    \\  \hline
        FG-CLIP$^{\dagger}$ (1.6B) & ViT-B/16 & $\surd$ & $\surd$ & $\times$  & 30.3 & 57.4 & 28.6 & 56.7 & 61.2 & 83.1 & 49.7 & 75.5 \\ \hline
		\rowcolor{lightgray} CLIP & ViT-B/16 & $\times$ & $\times$ & $\times$ &  18.6 & 40.6 &  25.5 & 43.1 & 41.4 & 63.6 & 30.6 & 53.8  \\
		CLIP-g & ViT-B/16 & $\times$ & $\times$ & $\times$  & 20.1 & 41.5 & 26.9  & 44.1 & 44.7 & 65.8 & 33.4 & 55.6 \\ 
		RegionCLIP & ViT-B/16 & $\surd$ & $\surd$ & $\surd$  &  27.9 & 54.4 & 34.2 & 54.3 & 59.4 & 80.9 & 47.4 & 73.2  \\  
		FineCLIP   & ViT-B/16  & $\surd$ & $\surd$ & $\times$  & 27.0 & 53.2 & 33.1  & 53.0 & 57.7 & 80.3 & 48.0 & 73.2  \\  
		DenseVLM   & ViT-B/16  & $\times$ & $\times$ & $\surd$  & 18.5 & 45.2 & 25.7  & 47.4 & 34.9 & 55.8 & 31.8 & 47.2 \\
        DenseVLM$^{\star}$  & ViT-B/16  & $\times$ & $\times$ & $\surd$  & - & - & -  & - & 60.1 & 79.9 & 49.4 & 62.4 \\ 
		CLIPSelf  & ViT-B/16  & $\surd$ & $\times$ & $\times$  & 27.9 & 55.3 & 33.6  & 53.2 & 60.9 & \textbf{81.0} & 49.0 & 73.2  \\  
        CLIM  & ViT-B/16  & $\times$ & $\times$ & $\times$  &  25.7 & 49.6 &  30.9 & 49.8 & 53.4 & 73.9 & 45.3 & 58.4 \\ 
		SFF-CLIP(Ours) & ViT-B/16 & $\times$ & $\times$ & $\times$  & \textbf{29.4} & \textbf{55.4} & \textbf{34.3}  & \textbf{54.5} & \textbf{62.2} & 80.8 & \textbf{52.1} & \textbf{73.6}  \\  \hline \hline
		\rowcolor{lightgray} CLIP & ViT-L/14 & - & - & -  & 31.2 & 56.8 &  41.2 & 62.2 & 58.1 & 78.9  & 49.8 & 72.6  \\
		CLIP-g  & ViT-L/14 & $\times$ & $\times$ & $\times$  & 33.0 & 59.2 & 44.2  & 64.5 & 60.3 & 80.9 & 54.3 & 71.2  \\
		SFF-CLIP(Ours) & ViT-L/14 & $\times$ & $\times$ & $\times$  & \textbf{36.9} & \textbf{66.7} &  \textbf{51.0} & \textbf{70.8} & \textbf{75.2} & \textbf{91.3} & \textbf{67.3} & \textbf{79.4}  \\ 	
		\hline 
		
	\end{tabular}
\vspace{-0.2cm}
\end{table*}

\myparagraph{Zero-shot region classification} To evaluate the dense representation ability, we use the mean accuracy (mAcc) of classifying region boxes and panoptic masks for ``things'' annotated in the ADE20K panoptic \cite{zhou2017scene} val set and COCO panoptic \cite{lin2014microsoft} val2017 set.
To extract region-level features, RoI or mask pooling are used to extract the region box or mask embeddings from the image dense feature maps. The classification is then performed by selecting the highest score when matching with the text embeddings of the classes.

The results are shown in Tab.~\ref{tab:zero-shot}. Compared with the CLIP base model,   
CLIP-g can slightly increase the zero-shot classification performance, due to additional iterations of image-level training. 
However, SFF-CLIP, which uses both the global contrastive learning and fine-grained matching, obtains significant improvements in accuracy on region classification, for both boxes and masks and for both ViT-B/16 and ViT-L/14 architectures. 
Compared with other region-aware fine-tuning methods,  
our proposed SFF-CLIP also achieves outstanding performances.
In contrast to other methods that require pre-preparing region proposals \cite{zhong2022regionclip,jing2024fineclip,wu2023clipself} and region labels \cite{zhong2022regionclip, jing2024fineclip}, or using pre-defined categories \cite{zhong2022regionclip, li2025densevlm}, 
our method can be flexibly applied with just image-text pairs as training data, which is the same data source as the original CLIP pre-training, resulting in better performance.
Furthermore, in contrast to the mosaicking-based CLIM \cite{wu2024clim}, our region-phrase pairs from self-annotation provide more effective fine-grained supervision than mosaicking images as pseudo regions, as evidenced by SFF-CLIP outperforming CLIM on all metrics.
These superior performances shows the effectiveness of our self-annotated fine-grained alignment approach, which even exceeds these methods that need to carefully prepare region annotations, and achieve comparable performances with the FG-CLIP \cite{xiefg2025fgclip}, which is a fine-grained model trained with large-scale region annotated data (1.6B).

\myparagraph{Open-vocabulary object detection} 
Following the previous related works \cite{wu2023clipself, jing2024fineclip, li2025densevlm}, we build open-vocabulary object detectors based on the F-ViT \cite{wu2023clipself} architecture, which is a two-stage detector using a frozen 
CLIP-based ViT as the backbone. In our experiment, each fine-grained fine-tuned CLIP encoder from the various methods is adopted to initialize the backbone. The OVD models are then trained on the OV-COCO benchmark \cite{chen2015microsoft}, and 
we use AdamW optimizer with batch size of 64, learning rate of 1e-4, and weight decay of 0.1.
For evaluation, we report box AP (average precision) at IoU (Intersection over Union) of base, novel and all categories as with previous works \cite{zhong2022regionclip,kuo2023fvlm,wu2023cora,kim2023region,wu2023clipself}.

The results are presented in Tab.~\ref{tab:app_ovd}. F-ViT is the baseline that initializes the detector backbone with the original pre-trained CLIP.
With the ViT-B/16 backbone, adopting just CLIP-g results in similar performance to the baseline CLIP. In contrast, SFF-CLIP  
significantly improves the OVD results, especially on the novel categories (over $9\%$). Since the base categories have explicit annotated bounding boxes and labels during OVD training, the performance on the unseen novel categories better illustrates the fine-grained understanding ability brought by the CLIP encoder.
Finally, we conduct  experiments with ViT-L/14 
and further improve the OVD performance on the novel categories by a large extent. We also provide comparisons with related works using ViT-L/14 in the Supplemental, as well as the OVD results with the LVIS benchmark \cite{gupta2019lvis}.
Compared with the existing OVD methods, which mostly rely on a ResNet-based encoder or modified ViT encoder and require pre-training on large-scale prepared data with extra region information, our method achieves superior or comparable performance by fine-grained training on the easily-obtained image-text pairs. 

\begin{figure*}[t]
	\begin{minipage}{.47 \linewidth}
    	\captionof{table}{Results for open-vocabulary object detection on MS COCO val set. F-ViT is the two-stage detector baseline built on the frozen original CLIP ViT,
    		and ``$+$'' means the ViT backbone is initialized with a fine-tuned model based on the corresponding method. 
            }
    	\label{tab:app_ovd}
    	\centering
    	\begin{tabular}{@{}lc|lll@{}}
    		\hline
    		Method  & Backbone & AP$_{50}^{novel}$ & AP$_{50}^{base}$ & AP$_{50}^{all}$    \\  \hline
    		OV-RCNN \cite{zareian2021open}  & ResNet50 & 17.5 & 41.0 & 34.9 \\
    		Detic \cite{zhou2022detecting} & ResNet50 & 27.8 & 51.1 & 45.0 \\
    		VLDet \cite{lin2022learning} & ResNet50 & 32.0 & 50.6 & 45.8 \\
    		F-VLM \cite{kuo2023fvlm} & ResNet50 & 28.0 & - & 39.6 \\
    		CORA \cite{wu2023cora} & ResNet50 & 35.1 & 35.5 & 35.4 \\
            SPARC \cite{bica2024improving} & ViT-B/16 & - & - & 39.4 \\
            FG-CLIP \cite{xiefg2025fgclip} & ViT-B/16 & 35.1 & 51.7 & 47.7 \\
            SigLIP2 \cite{tschannen2025siglip} & ViT-B/16 & - & - & 42.8 \\
    		RO-ViT \cite{kim2023region} & ViT-B/16 & 30.2 & - & 41.5 \\
    		RO-ViT \cite{kim2023region} & ViT-L/16 & 33.0 & - & 47.7 \\ \hline
    		
    		F-ViT  & ViT-B/16 & 19.4 & 43.3 & 37.0 \\
    		\hspace{4pt}+CLIP-g   & ViT-B/16 & 20.1 & 43.8 & 37.6 \\
    		\hspace{4pt}+RegionCLIP   & ViT-B/16 & 27.6  & 44.3 & 39.9\\
    		\hspace{4pt}+DenseVLM  & ViT-B/16 & 17.1 & 42.8 & 36.1 \\
            \hspace{4pt}+CLIM  & ViT-B/16 & 24.1  & 43.8 & 38.5 \\
    		\hspace{4pt}+CLIPSelf  & ViT-B/16 & 25.2 & 42.2 & 37.7 \\
    		\hspace{4pt}+FineCLIP  & ViT-B/16 & 27.2 & 46.0 & 41.1 \\
    		\hspace{4pt}+SFF-CLIP & ViT-B/16 & \textbf{28.8} & \textbf{46.4} & \textbf{41.8} \\ \hline
    		F-ViT  & ViT-L/14 & 28.3 & 52.5 & 46.2 \\
    		\hspace{4pt}+CLIP-g  & ViT-L/14 & 29.2 & 57.5 & 50.1 \\
    		\hspace{4pt}+SFF-CLIP  & ViT-L/14 & \textbf{37.4}  & \textbf{57.4} & \textbf{52.1} \\ 
    		\hline		
    		
    	\end{tabular}
    \end{minipage}
    \hspace{0.5cm}
    \begin{minipage}{.45 \linewidth}
	\captionof{table}{Results of open-vocabulary semantic segmentation on ADE20k-847 \cite{zhou2017scene}, Pascal VOC \cite{pascal-voc-2012}, and Pascal Context \cite{mottaghi2014role}. CatSeg is the segmentation baseline with the original CLIP ViT-B/16 as the backbone,
		and ``$+$'' means the ViT backbone is initialized with the fine-tuned model based on the corresponding method. 
        }
    \vspace{3pt}
	\label{tab:app_seg}
	\centering
	\begin{tabular}{@{}l|cc|cc@{}}
		\hline
		  &   \multicolumn{2}{c|}{VOC-20} &   \multicolumn{2}{c}{PC-59} \\
		Method  & mIoU  & pACC & mIoU  & pACC \\ \hline
		CatSeg    & 63.2  & 88.9 & 44.2  & 71.4 \\
		\hspace{7pt}+CLIP-g     & 73.8   & 92.1 & 50.1   & 74.4 \\
		\hspace{7pt}+RegionCLIP   & 78.4   & 93.9 & 54.9   & 76.2 \\
		\hspace{7pt}+DenseVLM   & 75.8  & 93.3 & 54.5  & 77.3\\
        \hspace{7pt}+CLIM  & 70.8  & 90.1 & 49.3 & 72.9 \\
		\hspace{7pt}+CLIPSelf & 75.1   & 92.3 & 51.6  & 75.6 \\
		\hspace{7pt}+FineCLIP    & 73.0 & 92.0 & 50.1  & 74.5 \\
		\hspace{7pt}+SFF-CLIP & \textbf{79.7}  & \textbf{94.2}  &  \textbf{56.9}  & \textbf{78.5} \\
		\hline	\hline

        & \multicolumn{2}{c|}{PC-459} & \multicolumn{2}{c}{A-847}\\
        Method & mIoU  & pACC & mIoU   & pACC \\  \hline
        CatSeg   & 8.7  & 38.2 & 5.8  & 25.0\\
        \hspace{7pt}+CLIP-g & 8.6  & 37.0 & 6.4   & 30.3\\
        \hspace{7pt}+RegionCLIP & 15.9  & 62.8 & 10.0   & 49.5\\
        \hspace{7pt}+DenseVLM  & 13.4  & 47.4 & 9.0  & 35.6\\
        \hspace{7pt}+CLIM   & 12.0 & 55.2 & 7.3 & 42.8 \\
        \hspace{7pt}+CLIPSelf  & 13.0  & 49.2 & 9.3  & 34.7\\
        \hspace{7pt}+FineCLIP & 9.3  & 37.9 & 6.7 & 28.7\\
        \hspace{7pt}+SFF-CLIP & \textbf{16.3}  & \textbf{65.3}  & \textbf{10.2}  & \textbf{58.4}\\
		\hline
		
	\end{tabular}        
    \end{minipage} 
\end{figure*}

\myparagraph{Open-vocabulary semantic segmentation} 
We next explore the performance of applying fine-grained fine-tuned models to open-vocabulary semantic segmentation. Following the previous works \cite{wu2023clipself, jing2024fineclip, li2025densevlm}, we adopt the CatSeg \cite{cho2024cat} as the segmentation architecture, which uses the dense image feature from CLIP ViT as the backbone, with the CLIP models fine-tuned by all compared methods using the same input image resolution 384x384. 
After replacing the backbone of CatSeg with the original CLIP model, and each region-aware model, the following segmentation experiments are conducted with training on COCO stuff \cite{caesar2018coco}, and evaluation on ADE20k \cite{zhou2017scene}, PASCAL VOC \cite{pascal-voc-2012}, and PASCAL Context \cite{mottaghi2014role} dataset using mean IoU (mIoU), and pixel Accuracy (pACC).  
As shown in Tab.~\ref{tab:app_seg}, SFF-CLIP comprehensively improves the performance of CatSeg on various datasets across different evaluation metrics, and surpasses the enhancements provided by RegionCLIP, CLIPSelf, DenseVLM, FineCLIP, and CLIM. 

Overall, the significant improvements on both OVD and OVS demonstrate that our method effectively boosts the fine-grained understanding ability of the CLIP model in downstream tasks.

\begin{figure*}[t]
	\centering
	\vspace{-0.1cm}
	\begin{center}
		\includegraphics[width=0.9 \textwidth]{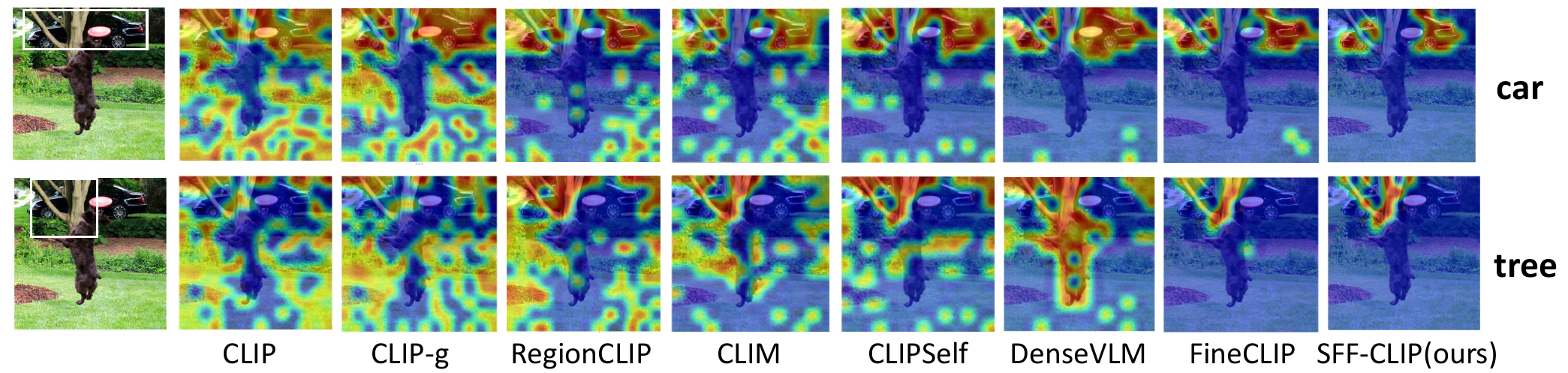}
	\end{center}
	\caption{Visualization of the fine-grained representations using similarity maps between the image dense feature and the text features for different objects (``car'' and ``tree'').
	}
	\vspace{-0.3cm}
	\label{fig:feat_sim}
\end{figure*}

\myparagraph{Qualitative comparison by visualization} In Fig.~\ref{fig:feat_sim}, we provide an example visualization to qualitatively compare the fine-grained understanding ability of models under different alignment methods. 
Cosine similarities are calculated between the embeddings on each position of the image's dense features and the text features for ``car'' and ``tree'', and visualized as similarity maps.  
Without fine-grained alignment, the similarity maps of vanilla CLIP and CLIP-g exhibit weak localization ability (i.e., region-specific attention), where many irrelevant locations have high similarities with the object text. 
For SFF-CLIP, the text can be matched with the specific regions more accurately than other methods. In particular many of the spurious matches to the background have been removed compared to other methods, which 
demonstrates the effectiveness of our proposed method to significantly improve the fine-grained representation ability of CLIP model.

\begin{table*}[t]
	\caption{Comparison of image-level representation by a zero-shot retrieval task using Flicker30k. 
		The gray row is the baseline CLIP before fine-grained fine-tuning, and $^{\dagger}$ represents the results from fine-grained models pre-trained by large-scale data, shown in the brackets.
        }
	\label{tab:comp}
	\centering
	\vspace{-0.2cm}    
	\begin{tabular}{lc|ccc|ccc@{}}
		\hline
		& &  \multicolumn{6}{c}{Flickr30k}  \\
		&  &  \multicolumn{3}{c|}{text-to-image} & \multicolumn{3}{c}{image-to-text} \\
		Method & Model   & R@1 & R@5 & R@10 & R@1 & R@5 & R@10    \\  \hline
        FG-CLIP$^{\dagger}$ (1.6B) & ViT-B/16  & 74.7 & 92.2 & 95.6  & 88.6 & 98.1 & 99.3  \\
        SPARC$^{\dagger}$ (3.2B) & ViT-B/16  & 72.0 & 91.2 & 94.9  & 84.4 & 97.6 & 98.7 \\ 
        SigLIP2$^{\dagger}$ (10B) & ViT-B/16  & 80.7 & - &  - & 93.0 & - & - \\ \hline
		\rowcolor{lightgray} CLIP & ViT-B/16  & 73.6 & 90.9 &  94.8 & 88.6 & 97.1 & 99.1 \\
		CLIP-g & ViT-B/16   &  76.9 & 93.0 & 95.9 & 90.8 & \textbf{98.8} & \textbf{99.4}\\
		RegionCLIP & ViT-B/16 & 70.1  & 90.2 & 94.3 & 81.1 & 94.8 & 97.0 \\  
		FineCLIP & ViT-B/16  & 67.0 & 88.5 & 93.6 & 79.7 & 96.3 & 97.9  \\  
		DenseVLM  & ViT-B/16  & 32.4 & 56.7 & 67.4 & 17.8 & 34.0 & 45.9 \\ 
		CLIPSelf  & ViT-B/16  & 52.3 & 78.5 & 86.0 & 49.9 & 77.2 & 85.4\\  
        CLIM  & ViT-B/16 & 71.2 & 90.3 & 94.4 & 84.1 & 97.3 & 98.5 \\
		SFF-CLIP(Ours) & ViT-B/16  & \textbf{77.4} & \textbf{93.1} & \textbf{ 95.9 } & \textbf{90.8} & 98.5 & 99.3 \\  \hline \hline
		\rowcolor{lightgray} CLIP & ViT-L/14 & 78.8 & 94.0 & 96.8  & 90.4 & 98.8 & 99.4  \\
		CLIP-g  & ViT-L/14 & \textbf{82.8} & 95.9 & \textbf{97.9}  & 93.7  & 99.2 & 99.8 \\
		SFF-CLIP(Ours) & ViT-L/14 & 82.6 & \textbf{95.9} & 97.8 & \textbf{94.0} & \textbf{99.3} & \textbf{99.8} \\ \hline	
		
	\end{tabular}
\end{table*} 

\subsection{Evaluation of image-level representation} 
\vspace{-0.2cm}
The previous works 
rarely considered how their fine-tuning affects the whole image-text matching, which is the original ability of a vision-language model (VLM). 
However, preserving the original image-level representation is important in order to support multiple tasks from the same feature backbone model.
To explore if there are any negative effects to image-level representation when improving fine-grained representations, we conduct an image-text retrieval evaluation with the Flickr30k \cite{plummer2015flickr30k} validation set, and report the recall accuracy of image-to-text and text-to-image on Top$@\{1, 5, 10\}$ matching. 
As seen in Tab.~\ref{tab:comp}, with just realizing image-level contrastive learning, the baseline CLIP-g can keep or improve the performance of the original CLIP by continuing the training on the new image-text pairs data from MS COCO. 
Our SFF-CLIP successfully preserves the image-level retrieval performance, which is comparable to or even better than CLIP-g.
In contrast, other region alignment methods all cause degradation to the image-text retrieval performances compared with the CLIP-g. 
In particular, CLIPSelf and DenseVLM have largely reduced their abilities of global image-level representation.  
Finally, the similar results between SFF-CLIP and CLIP-g with ViT-L/14 further demonstrate the effectiveness of our method in maintaining the global representation and stable image-level matching. 

\begin{figure*}[t]
	\vspace{-0.1cm}	 
	\begin{minipage}{.31 \linewidth}
		\captionof{table}{Ablation study of phrase matching weight $w_{t}$.}
		\label{tab:abl_phrase_w}
		\centering
		\setlength\tabcolsep{1.5 pt}   
		\begin{tabular}{lccc}
			\hline
			Method   & & Boxes & Masks   \\  \hline
			\rowcolor{lightgray}	CLIP & - & 41.4  & 30.6   \\  \hline
            \multirow{2}*{SFF-CLIP} & w/o $w_{t}$ & 39.5  & 28.5   \\
			& w/ $w_{t}$  & \textbf{57.8}  & \textbf{50.1}   \\     
			\hline 
			
		\end{tabular}
		\vspace{-0.2cm}	
	\end{minipage}
	\hspace{0.03cm}
	\begin{minipage}{.3 \linewidth}
		\captionof{table}{Ablation study of the momentum model $\hat{M}$.}
		\label{tab:abl_momentum}
		\centering
		\setlength\tabcolsep{1.5 pt}    	
    		\begin{tabular}{lcc@{}}
    			\hline
    			Method   & T2I   & I2T  \\  \hline
                \rowcolor{lightgray}	CLIP-g w/o $\hat{M}$ &  76.2  & 90.5    \\
    			\rowcolor{lightgray}	CLIP-g w/ $\hat{M}$ &  76.9  &  90.8   \\  \hline
    			SSF-CLIP w/o $\hat{M}$ & 74.9   & 88.8  \\
    			SSF-CLIP w/ $\hat{M}$ &\textbf{77.4}   &\textbf{ 90.8}   \\ 
    			\hline 
		\end{tabular}
		\vspace{-0.2cm}	
	\end{minipage}
	\hspace{0.03cm}
	\begin{minipage}{.35\linewidth}
		\captionof{table}{Ablation study of number of extracted phrase $n$.}
		\label{tab:abl_phrase_num}
		\centering
		\setlength\tabcolsep{1.5 pt}    
		\begin{tabular}{lcccc}
			\hline
			Method   & N & T/epoch & Boxes  & Masks   \\  \hline
			CLIP-g & 0 & 51min & 44.7 & 33.4   \\  \hline
			\multirow{3}*{SFF-CLIP} & 1 & 53min & 60.7  & 49.3   \\
			& 3 & 58min  & 61.8  & 51.7    \\    
			& 5  & 1h & \textbf{62.2}  & \textbf{52.1}   \\   
			\hline 
			
		\end{tabular}
		\vspace{-0.2cm}	
	\end{minipage}
\vspace{-0.2cm}	
\end{figure*}

\subsection{Ablation Studies}
\vspace{-0.2cm}
In this section, we conduct the ablation studies showing the influence of various components in our fine-grained fine-tuning method.

\label{sec:ablation}

\myparagraph{Phrase matching $w_{t}$} To show the function of the phrase-matching weight $w_{t}$ in the fine-grained matching loss (Eq.~\ref{eq:local_loss}), we conduct an ablation study with and without the weight on zero-shot classification performance, as the Top-1 accuracy presented in Tab.~\ref{tab:abl_phrase_w}. Without $w_{t}$, the fine-grained alignment model degrades significantly, with results even worse than the original CLIP. This demonstrates that many phrases extracted from the caption are not matched well with the image, and could cause large disturbances to the region-phrase alignment. Therefore, after introducing the weights to represent the degree of matching for each phrase into the loss, the model can better focus on the learning of well-aligned samples during the fine-tuning, and produce significant performance improvements.

\myparagraph{Momentum model for image-level representation} We conduct an ablation study to investigate the function of using momentum model to generate image and text matching weights in the contrastive learning (Eq.~\ref{eq:global_loss}). From the text-to-image (T2I) and image-to-text (I2T) results shown in Tab.~\ref{tab:abl_momentum}, the image-text retrieval evaluation results are well kept when adopting the momentum model, but decrease when using the pure contrastive loss without the weights from the momentum model.

\myparagraph{Maximum number of phrases}
Since the text-specific heat maps can be calculated in batch for one phrase in each sentence, the number of loops in obtaining all heat maps is the same as the maximum number of phrases $N$. Therefore, we conduct the ablation study with $N=\{0,1,3,5\}$ to observe the influence on the fine-grained alignment's effectiveness and training efficiency. From  Tab.~\ref{tab:abl_phrase_num}, the zero-shot classification Top-1 accuracy is similar when $N=3$ and $N=5$, which is better than $N=1$, while much higher than no region-phrase matching (CLIP-g). Thus, 
$N={3,5}$ can basically cover the effective concepts in the training data. For the training efficiency, due to the high speed of the adopted heat-map generation method, the influence of $N$ on the training time cost is not large. 

\begin{figure}[t]
	\centering
	\begin{center}
		\includegraphics[width=0.98 \textwidth]{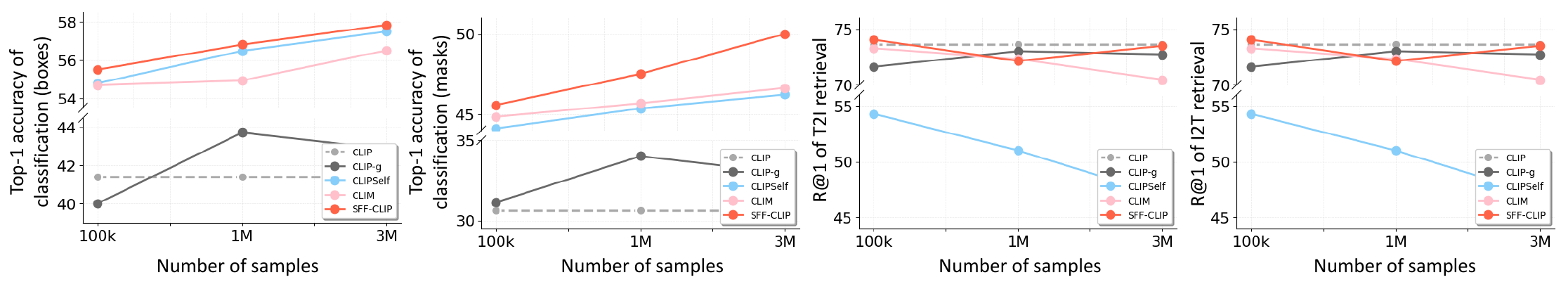}
	\end{center}
	\caption{Ablation study on training data in three different scales. }
	\label{fig:data_scale}
	\vspace{-0.2cm}
\end{figure} 

\myparagraph{Training data scale}
To investigate the impact of data scale on model performance, we randomly sample three trainsets from CC3M \cite{changpinyo2021conceptual} with different sizes: 100K, 1M, and 3M samples. 
We fine-tune the CLIP model with our SSF-CLIP, and compare with CLIP-g, CLIM and CLIPSelf, where CLIPSelf uses image patches instead of preparing region proposals.
We report the Top-1 accuracy of zero-shot classification with boxes or masks for fine-grained understanding evaluation and the Top-1 recall rate of T2I and I2T retrieval results for image-level representation evaluation. In Figure~\ref{fig:data_scale}, we present the performance curves on the four evaluations as the number of samples scales up. In terms of the fine-grained understanding, SSF-CLIP surpasses other methods, and the performance continues to grow as the dataset size increases, which shows promising scalability. For the image-level representation, SSF-CLIP maintains similar retrieval performances as CLIP-g, while in contrast,  CLIPSelf and CLIM become worse when the training data increases.

\section{Conclusion}
\label{sec:conclusion}
In this paper, we propose SFF-CLIP (Self-annotated Fine-grained Fine-tuning for CLIP), a novel framework designed to advance the fine-grained representation ability while maintaining the global visual-semantic consistency.
Our framework only requires image-text pairs as inputs, avoiding the process of generating region proposals and region annotations (eithetasksnually or via detectors) required by previous works.
We propose fine-grained matching with the help of a run-time region-phrase annotation and alignment scheme, which extracts concept phrases from the input sentence and dynamically matches them with image dense features with the help of text-specific heat maps.
A momentum model is adopted to support the preservation of CLIP's original image-level representation ability.
We validate the fine-grained dense representation of SFF-CLIP 
on zero-shot classification task, and down-stream dense prediction tasks, producing consistently significant performance improvements to the baseline model, as well as superior results compared to state-of-the-art works.  
Meanwhile, the performance on image-level tasks is preserved from the original CLIP model. 
Training with just image-text pairs, SFF-CLIP provides a flexible and effective solution for improving fine-grained dense representations of vision-language models like CLIP, while also eliminating the requirements of extra training data annotations, such as region proposals and region annotations, as well as dependence on pre-defined categories.

\clearpage

\section*{Acknowledgments}
This was was supported in part by......

\appendix
\section{Comparisons on ViT-L/14}
We compare with related fine-grained fine-tuning methods, including RegionCLIP \cite{zhong2022regionclip}, CLIPSelf \cite{wu2023clipself}, FineCLIP \cite{jing2024fineclip}, DenseVLM \cite{li2025densevlm}, and CLIM \cite{wu2024clim}, on ViT-L/14. For fairness, all compared methods are reproduced with their open-sourced codes to fine-tune the CLIP pre-trained model based on the MS COCO train 2017 set \cite{lin2014microsoft} with the same input image size of 336x336. 

\begin{table*}[h]
	\caption{Comparison of fine-grained dense representations via zero-shot classification on the ADE20K panoptic \cite{zhou2017scene} val set and COCO panoptic \cite{lin2014microsoft} val2017 set. We report the Top-1 and Top-5 mean accuracy on both object bounding boxes and panoptic masks. 
	The gray row is the baseline CLIP before fine-grained alignment fine-tuning. 
    Our method does not require preparation of region proposals (R.P.), region labels (R.L.), or predefined categories (P.C.) on the training data.
    }
	\label{tab:supp-zero-shot}
	\centering
	\begin{tabular}{lc|ccc|cc|cc||cc|cc@{}}
		\hline
		& &  &  &   & \multicolumn{4}{c||}{ADE20k} &  \multicolumn{4}{c}{MS COCO} \\
		&  &   &   &   & \multicolumn{2}{c}{Boxes} & \multicolumn{2}{c||}{Masks} & \multicolumn{2}{c}{Boxes} & \multicolumn{2}{c}{Masks}\\
		Method & Model  & R.P. &  R.L. & P.C.  & Top1 & Top5 & Top1 & Top5  & Top1 & Top5 & Top1 & Top5    \\  \hline
		\rowcolor{lightgray} CLIP & ViT-L/14 & - & - & -  & 31.2 & 56.8 &  41.2 & 62.2 & 58.1 & 78.9  & 49.8 & 72.6  \\
		CLIP-g  & ViT-L/14 & $\times$ & $\times$ & $\times$    & 33.0 & 59.2 & 44.2  & 64.5 & 60.3 & 80.9 & 54.3 & 71.2  \\
		RegionCLIP & ViT-L/14  & $\surd$ & $\surd$ & $\surd$   & 30.7  & 57.0 & 42.9 &  66.3 & 66.1 & 88.0 & 56.4 & 76.8  \\  
		FineCLIP   & ViT-L/14  & $\surd$ & $\surd$ & $\times$  & 32.0  & 58.6 & 46.1 & 67.8 & 68.2 & 87.8 & 60.5 & 77.0 \\  
		CLIPSelf   & ViT-L/14  & $\surd$ & $\times$ & $\times$ &  36.1 & 66.4 & 49.6 & 70.1 & 72.0 & 90.1 & 65.3 & 78.3  \\  
        CLIM  & ViT-L/14  & $\times$ & $\times$ & $\times$     & 30.4  & 55.6 & 41.7 & 63.8 & 60.2 & 81.5 & 58.6 & 74.0  \\ 		
        SFF-CLIP(Ours) & ViT-L/14 & $\times$ & $\times$ & $\times$  & \textbf{36.9} & \textbf{66.7} &  \textbf{51.0} & \textbf{70.8} & \textbf{75.2} & \textbf{91.3} & \textbf{67.3} & \textbf{79.4}  \\ 
		\hline 
		
	\end{tabular}
\end{table*}

\subsection{Results of zero-shot region classification}
We compare the fine-grained dense representation on ViT-L/14 with related fine-grained fine-tuning methods using the mean accuraccy (mACC) of classifying region boxes and panoptic masks for ``things''. The results on ADE20K panoptic \cite{zhou2017scene} val set and COCO panoptic \cite{lin2014microsoft} val2017 set are shown in Tab.~\ref{tab:supp-zero-shot}. Compared to other region-aware fine-tuning methods that require pre-preparing extra region information, our proposed SFF-CLIP achieves superior performance with the ViT-L/14 architecture, further demonstrating the effectiveness of our self-annotated fine-grained alignment on advancing the region-aware representation ability of CLIP. 

\subsection{Results of OVD with OV-COCO benchmark}
 We provide the results of OVD performance on MS COCO val set with ViT-L/14 in Tab.~\ref{tab:sup_ovd_vitl}. SFF-CLIP significantly improves the OVD results on the unseen novel categories (9.1\%), while surpassing other fine-grained fine-tuning methods. 

\begin{table}[h]
    \setlength\tabcolsep{4pt}
    \centering
    \caption{Results for open-vocabulary object detection on MS COCO val set. F-ViT is the two-stage detector baseline built on the frozen original CLIP ViT, and ``$+$'' means the ViT backbone is initialized with a fine-tuned model based on the corresponding method.}
    \begin{tabular}{lc|ccc}
        \hline
        Method  & Backbone & AP$_{50}^{novel}$ &  AP$_{50}^{base}$  &  AP$_{50}^{all}$ \\ \hline 
          F-ViT & ViT-L/14 & 28.3 & 52.5 & 46.2 \\
        \hspace{4pt}+CLIP-g & ViT-L/14 & 29.2 & \textbf{57.5} & 50.1 \\
        \hspace{4pt}+RegionCLIP & ViT-L/14 & 36.9 & 52.8  & 48.7 \\
        \hspace{4pt}+FineCLIP  & ViT-L/14 & 37.2 & 54.3 & 49.8 \\
        \hspace{4pt}+CLIPSelf  & ViT-L/14 & 30.0  &  53.8  & 47.4 \\ 
        \hspace{4pt}+SFF-CLIP(Ours)  & ViT-L/14 & \textbf{37.4} & 57.4  & \textbf{52.1} \\ 
        \hline 
    \end{tabular}
    \label{tab:sup_ovd_vitl}
\end{table}

\subsection{Results of image-text retrieval}
Moreover, the comparison results on the image-text retrieval task shown in Tab.~\ref{tab:sup-image-level} also indicate that our SSF-CLIP can successfully maintain the global representation and stable image-level matching with ViT-L/14. In contrast, other fine-grained fine-tuning methods all cause degradation of image-text retrieval performances, especially the region-based distillation method CLIPSelf. 

\begin{table*}[h]
	\caption{Comparison of image-level representation by a zero-shot retrieval task using Flicker30k \cite{plummer2015flickr30k}. 
		The gray row is the baseline CLIP before fine-grained fine-tuning, and $^{\dagger}$ represents the results from fine-grained models pre-trained by large-scale data, shown in the brackets.
        }
	\label{tab:sup-image-level}
	\centering
	\vspace{-0.2cm}    
	\begin{tabular}{lc|ccc|ccc@{}}
		\hline
		& &  \multicolumn{6}{c}{Flickr30k}  \\
		&  &  \multicolumn{3}{c|}{text-to-image} & \multicolumn{3}{c}{image-to-text} \\
		Method & Model   & R@1 & R@5 & R@10 & R@1 & R@5 & R@10    \\  \hline
		\rowcolor{lightgray} CLIP & ViT-L/14 & 78.8 & 94.0 & 96.8  & 90.4 & 98.8 & 99.4  \\
		CLIP-g  & ViT-L/14 & \textbf{82.8} & 95.9 & \textbf{97.9}  & 93.7  & 99.2 & 99.8 \\
        RegionCLIP & ViT-L/14 &  75.8 & 92.5 & 95.9 & 85.8 & 96.9 & 98.8 \\  
		FineCLIP & ViT-L/14 & 70.9  & 90.3 & 94.4 & 80.4 & 96.5 & 98.8 \\
		CLIPSelf  & ViT-L/14  & 15.8  & 35.8 & 47.7 & 6.0 & 16.5 & 23.0 \\
        CLIM  & ViT-L/14 &  73.7 & 92.2 & 95.4 & 87.5 & 97.5 & 99.0 \\
		SFF-CLIP(Ours) & ViT-L/14 & 82.6 & \textbf{95.9} & 97.8 & \textbf{94.0} & \textbf{99.3} & \textbf{99.8} \\ \hline
	\end{tabular}
\end{table*}

\section{OVD results on OV-LVIS benchmark}
We build open-vocabulary object detectors based on F-ViT architecture using the fine-grained fine-tuned CLIP ViTs as backbones, and train the models for 48 epochs on the OV-LVIS \cite{gupta2019lvis} benchmark with the input image size of 224x224 for ViT-B/16. The evaluation results are shown in the Tab.~\ref{tab:sup_app_ovd} with AP for base categories (AP$_{c}$, AP$_{f}$), rare categories (AP$_{r}$), and all categories (AP) as comparison indicators. SSF-CLIP surpasses all other fine-grained fine-tuning methods on all categories, especially on the rare categories, which better illustrates the fine-grained understanding ability brought by the CLIP encoder.

\begin{table}[h]
    	\captionof{table}{Results for open-vocabulary object detection on OV-LVIS val set. F-ViT is the two-stage detector baseline built on the frozen original CLIP ViT, and ``$+$'' means the ViT backbone is initialized with a fine-tuned model based on the corresponding method. 
        }
    	\label{tab:sup_app_ovd}
    	\centering
    	\begin{tabular}{@{}lc|cccc@{}}
    		\hline
    		Method  & Backbone & AP & AP$_{r}$ & AP$_{c}$ &  AP$_{f}$   \\  \hline
    		F-ViT  & ViT-B/16 & 9.5 & 3.1 & 6.6 & 15.7 \\
    		\hspace{4pt}+CLIP-g   & ViT-B/16  & 10.7 & 5.1 & 7.4 & 16.1 \\
    		\hspace{4pt}+RegionCLIP   & ViT-B/16 & 10.4 & 5.8 & 7.2 & 16.1 \\
    		\hspace{4pt}+CLIPSelf  & ViT-B/16 & 9.0 & 3.6 & 5.8 & 14.9 \\
    		\hspace{4pt}+FineCLIP  & ViT-B/16 & 10.2 & 4.2 & 7.0 &  16.4 \\
    		\hspace{4pt}+SFF-CLIP(Ours) & ViT-B/16 & \textbf{11.1} & \textbf{7.5} & \textbf{7.6} & \textbf{16.7} \\ 
    		\hline		
    		
    	\end{tabular}
\end{table}

\section{Ablation study on different XAI methods}
\label{sec:abl_xai}
In Sec. 3.2, we calculate the text-specific heat maps region-phrase self-annotation based on explainable AI (XAI) methods, and specifically adopt Grad-ECLIP \cite{zhao2024gradient} as reference. 
To investigate the influence of the XAI method used for fine-grained alignment, we conduct an ablation study with other applicable visual explanation approaches that satisfy a high-speed and easy plug-in, including Grad-CAM \cite{selvaraju2017grad} and MaskCLIP \cite{zhou2022extract}. The Vision Transformer self-attention cannot be used since it is not text-specific.
The comparison of zero-shot classification performances in Table~\ref{tab:abl_xai} shows that with the heat maps from these two methods, the phrase-region alignment also obtains obvious performance improvements compared with just using global loss (CLIP-g), but there is still a gap compared with using Grad-ECLIP, which demonstrates that Grad-ECLIP is the most effective.

\begin{table}[h]
	\caption{Ablation study of using different visual explanation heat maps in SSF-CLIP.}
	\label{tab:abl_xai}
	\centering
	\begin{tabular}{lccccc}
		\hline
		&  & \multicolumn{2}{c}{Boxes} & \multicolumn{2}{c}{Masks}  \\
		Method   & Explanation Map & Top1 & Top5 & Top1 & Top5   \\  \hline
		\rowcolor{lightgray}	CLIP  &  -  & 41.4 & 63.6 & 30.6 & 53.8  \\  
		CLIP-g & - & 42.9 & 64.8 & 32.9 & 56.4  \\  \hline
		\multirow{3}*{SSF-CLIP} &  Grad-CAM & 54.2 & 74.7 & 46.5 & 69.8  \\
		&  MaskCLIP & 54.3 & 75.5 & 47.4 & 70.9  \\
		&  Grad-ECLIP &\textbf{57.8}& \textbf{78.6} & \textbf{50.1} & \textbf{72.9}  \\     
		\hline 
		
	\end{tabular}
	\vspace{-0.1cm}	
\end{table}

\section{Ablation study on input image sizes}
\label{sec:abl_imgsize}
To explore the impact of input image size on SSF-CLIP, we conduct the fine-grained alignment training and evaluation with four different image resolutions, including $224, 336, 480, 512$, on both region-level zero-shot classification and image-level retrieval. The results are shown in Table~\ref{tab:abl_imgsize}. As the resolution of input image is increased gradually from $224$ to $512$, the performance on region-level task is improved due to that more visual details can be provided with larger images. On the other hand, the performance on the image-text retrieval task is declined as the image becomes larger, when more number of patches leads to a increase of the complexity for the $[cls]$ token to integrate the global feature.

\begin{table}[h]
	\caption{Ablation study on input image sizes.}
	\label{tab:abl_imgsize}
	\centering
	\begin{tabular}{ccccc|cccc}
		\hline
		& \multicolumn{4}{c|}{zero-shot classification} & \multicolumn{4}{c}{image-text retrieval} \\
		Image & \multicolumn{2}{c}{Boxes} & \multicolumn{2}{c|}{Masks} & \multicolumn{2}{c}{text-to-image} & \multicolumn{2}{c}{image-to-text}  \\
		 Size  & Top1 & Top5 & Top1 & Top5 & R@1 & R@5 & R@1 & R@5  \\  \hline
		  224   & 62.2 & 80.8 & 52.1 & 73.6 & 77.4 & 93.1 & 90.8 & 98.5   \\ 
		  336   & 63.3 & 82.1 & 53.7 & 74.5 & 76.1 & 91.8 & 88.8 & 98.5   \\ 
		  480   & 65.5 & 86.2 & 59.1 & 81.1 & 70.5 & 88.9 & 85.5 & 96.3   \\ 
		  512   & 66.2 & 86.8 & 65.0 & 81.4 & 66.6 & 86.6 & 81.5 & 95.2  \\

		\hline 
		
	\end{tabular}
	\vspace{-0.2cm}	
\end{table}

\section{Visualization}
\begin{figure*}[th]
	\centering
	\begin{center}
		\includegraphics[width=0.96 \textwidth]{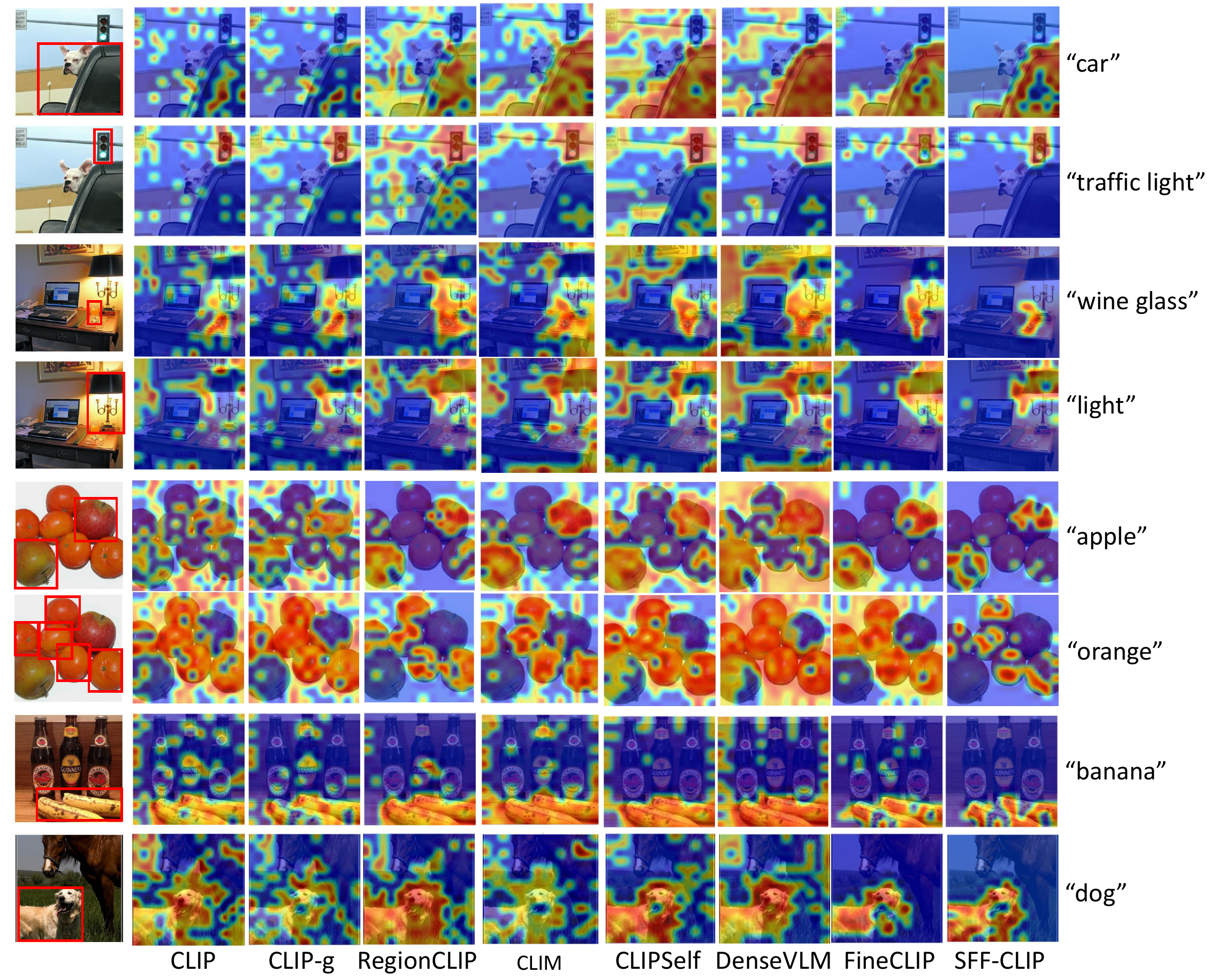}
	\end{center}
	\vspace{-0.3cm}
	\caption{Visualization of cosine similarity maps between text embeddings and the dense feature maps generated by CLIP, CLIP-g, RegionCLIP \cite{zhong2022regionclip}, CLIM \cite{wu2024clim}, CLIPSelf \cite{wu2023clipself}, DenseVLM \cite{li2025densevlm}, FineCLIP \cite{jing2024fineclip} and our SSF-CLIP. }
	\label{fig:sup_vis}
	\vspace{-0.3cm}
\end{figure*}

We present more visualizations of cosine similarity maps between text embeddings and the dense feature maps generated by different fine-grained alignment methods. The comparisons are shown in Figure~\ref{fig:sup_vis}.

\section{Baselines}
We use the following publicly available source code:
\begin{compactenum}
    \item CLIPSelf \cite{wu2023clipself} \& RegionCLIP \cite{zhong2022regionclip}: https://github.com/wusize/CLIPSelf
    
    \item CLIM \cite{wu2024clim}: https://github.com/wusize/CLIM
    
    \item FineCLIP \cite{jing2024fineclip}: https://github.com/Timsty1/FineCLIP
    \item DenseVLM \cite{li2025densevlm}: https://github.com/HVision-NKU/DenseVLM
\end{compactenum}

\section{Broader impact}
\label{sec:broader_impact}
Our work contribute to introduce SSF-CLIP, which boosts the fine-grained understanding ability of CLIP while maintains the global representation with eliminating the constrains from region annotation preparation. With the growing adoption of transformer as a unified architecture for both vision and language tasks, it is of great significant to enable the great generalization ability of CLIP ViT encoders in both image-level and dense prediction tasks. We are the first to utilize the XAI to produce self-annotated supervision. 
Eliminating the requirements of cumbersome region data preparation, the method is expected to support larger scale of data and promote improvements of models with further development.
To ensure a positive social impact, we conduct experiments using academic open-source datasets that do not involve personal privacy issues.

\section{Limitation}
\label{sec:limitation}
The SSF-CLIP is built upon an effective visual explanation (XAI) method for CLIP, the Grad-ECLIP \cite{zhao2024gradient}. The self-annotated region-phrase alignment will be influenced by the performance of heat maps generated by the adopted XAI method, as shown in \S\ref{sec:abl_xai}. Therefore, the generalization of our idea that utilizes XAI to boost the model itself can be limited by the development of corresponding XAI technique. On the other side, the further progress made in XAI area may push our approach to have a wider range of applications. 
Since our work provides a flexible and low-cost way to supplement or replace the manually labeling process, SSF-CLIP has the potential to further enhance other pretrained models, which would be explored in our future work.

\bibliographystyle{unsrt}  
\bibliography{references}

\end{document}